# Human Face as Human Single Identity

Spits Warnars
Department of Computing and Mathematics, Manchester Metropolitan University
John Dalton Building, Chester Street, Manchester M15GD, United Kingdom
s.warnars@mmu.ac.uk

**Abstract** :
Human face as a physical human recognition can be used as a unique identity for computer to recognize human by transforming human face with face algorithm as simple text number which can be primary key for human. Human face as single identity for human will be done by making a huge and large world centre human face database, where the human face around the world will be recorded from time to time and from generation to generation. Architecture database will be divided become human face image database which will save human face images and human face output code which will save human face output code as a transformation human face image with face algorithm. As an improvement the slightly and simple human face output code database will make human face searching process become more fast. Transaction with human face as a transaction without card can make human no need their card for the transaction and office automation and banking system as an example for implementation architecture. As an addition suspect human face database can be extended for fighting crime and terrorism by doing surveillance and searching suspect human face around the world.

## 1. Introduction

In daily live human recognize others by looking their faces. For the first time humans will be difficult to recognize new human face which have never been seen before. Slowly but sure people can recognize others by continuous recognition. In surveillance system human has been helping by camera technology such as CCTV or webcam in order to recognize other human for security surveillance. Computer technology as a part of surveillance tools has been extended as think like a human not just only to record the video picture but doing the surveillance by recognizing human face. Recently face recognition research have been doing as a part of computer vision research and many applications have been creating based on face recognition concept.

In daily transaction lives human have been helping by computer to recognize others by numbering human as their identity such as Passport number, Student number, Account number, employee number, National Insurance number, National Health number and so on. Man has many numbers which is mean man has many identities and having many identities for man can be susceptible for crime or deception. Single identity for man will help the system to recognize human, reduce crime or deception and improve security.

Human face as a physical recognition for human can be used as an identity for computer to recognize human no matter that lack of current face recognition. Face has several advantages over other biometric technology like finger, hand, voice, eye and signature based on number of evaluation factor such as enrolment, renewal, machine requirement and public perception [1]. Moreover face in video is more acceptable and collectable rather than other biometrics recognition like iris, fingerprint or even face in document [2].

In this paper we will deepen for how to make a system where human face as a single identity for human and what the right infrastructure for this single identity system. How the system can recognize the human face, store and manage it?

## 2. Human face as a primary key

In daily live human can be easy to recognize others if they already had the memory about other human face. In computer system the human memory which placed in human brain is a database which can save the data or transaction as a process for memorizing. It's clear that for building the system which can recognize human face will be needed the database as memorizing. Best structure database design for recognizing human face will influence the algorithm as a process and best algorithm will influence the best structure database and obviously will influence the better and faster human face computer system.

As computer can recognize human by numbering human as their identity then in most information system human has been recognized as a unique record in database which had been saved before. In order to make a system where can recognize human face and differ between each others, then the idea is to make a unique record in database for man as well. The idea is to transform human face as number which can be saved as primary key in database.

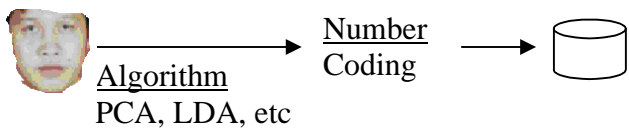

Figure 1. Human face numbering

Some of Face recognizing algorithm like PCA (Principal Component Analysis), IDA (Independent Component Analysis), LDA (Linier Discriminate Analysis) and others can be chose or extended to transform human face into number which can be saved as a unique record in database. A unique record which primary key as unique identity for human.

### 3. World Centre Human face database

In order to make single identity for human face as security improvement and for reducing crime and deception then a single and centre database must be created as well as a world centre human face. The centre large human face databases contain with human faces data from all the countries, from time to time and from generation to generation. Man not only be recognized in one country but can be recognized in other countries as one personal with one identity. Moreover Man can be recognized from time to time and from generation to generation. As addition genealogy system can be built significantly based on this database.

A single and large centre database will imply the needed for big resources and as a result the best structure database design and algorithm must be improved. Beside that for increasing the performance, database technology can be implemented such as distribution database, data warehouse, data mining and etc.

For performance reason database will be divided as:
1. Human face image
2. Human face output code

Human face image database will be recorded with human face images and because of saving image can make database become a huge database then this database just a link from slightly Human Face Output Code database. Possibly one record in Human Face Output Code database will have many records in Human Face Image database. This can be happened because human face appearance can change periodically as a result for ages, accident or anything which can change the appearance human face.

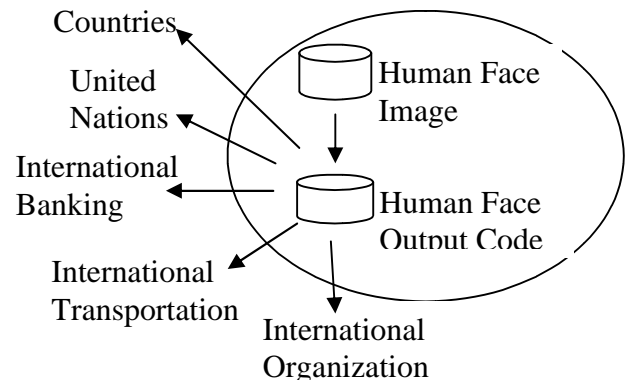

Figure 2. world Centre Human Face Database

Human Face Output Code database will be recorded with output code as a result from Face Algorithm. Data which recorded as a represent for human face code in appearance simple text will make database become slightly and for best performance searching purpose. This database will be used for searching human identity and become a link to Human Face Image database. Each record will represent as unique human identity and act as primary key. For security purpose encryption and decryption can be used for this simple text record.

Human Face Output Code become an interface to other systems, where external systems like United Nations, Countries, Banking system, International transportation and international organization can access the database. United Nations and international organization can get a real time data about human demography.

Countries will be helped by Centre Human Face Database where they can control human travelling to access their country and national security will be improved. People no need passport anymore to travelling and countries no need to issue a passport or a physical visa which issued on passport. Visa can be issued as electronic visa issued. Human face substitute passport as unique identification for human. In the future we do not need passport and visa for international travelling.

It will be better if all countries are aware to implement this technology. Developed countries can help developing countries for the implementation and as a reward developed countries will increase their national security. As a result system one single identity will improve security and accountability in international banking system, international transportation, and other international system.

For against the international criminal, there will be a special database which can record human which have criminal records, specifically for international criminal or terrorist activities. The most wanted people can be recognized by searching their face on every place which can record the activities as a video or image sequences.

In banking system international cash flow can be detected very early. Every people can be detected where they put their money, what the resources for their income and the purpose for their fund. Banking system can protect against the money laundry and using fund for crime purposing such as funding for terrorist organization. For international transportation will be easy to control human travelling, beside electronic visa is implemented and people no need passport anymore.

There are common thought that human obey the law when there is a significance law, punishment and surveillance. Human obey the law because they are surveyed, human are not going to contravene traffic lamp because there are camera surveillance and human are going to slow their car speed when there is camera surveillance. Absolutely human must need surveillance for living harmony in this world and as a result one single identity can decrease the private confidential, but as long as there are some protection rules, accountability and everything is under control then security will be improved.

For increasing the performance of database distribute database will be implemented where primary key will be duplicated in many external databases.

4. **Transaction without card**

For many transaction systems, the system can easily to recognize human when they bring card as personal identity. But when human forget or lost their card then they can't do their transaction easily, the system can't recognize them without their card.

Human face as an identity for transaction can make people no need to bring their card anymore, people never worry to lose or forget to bring their card as an identity. The System can be helped for recognizing human without need their card as long as they show their face. Man with one single identity can be improved and even for the system in the whole world and not only for the system in one country.

Using human face as an identity for transaction can be implemented in many sector like office automation, where the system can detect human face and give a message alert. For banking system people no need their ATM/Debit card anymore, they don't need to remember their password, security number or pin anymore, they just show their face to access cash machines, pay for goods and services at the retailer or have a dealing with internet banking.

In office automation there will be 4 kinds of databases
1. Face output code database
2. Face image database
3. Message database
4. Personal database

The same like the database architecture for world centre human face database, face image database will be captured with images of human face in many pose and appearance. Face output code database will be recorded with output code as a simple text as a result from face algorithm which is primary key. It is possible to connect these databases to world centre human face database as well.

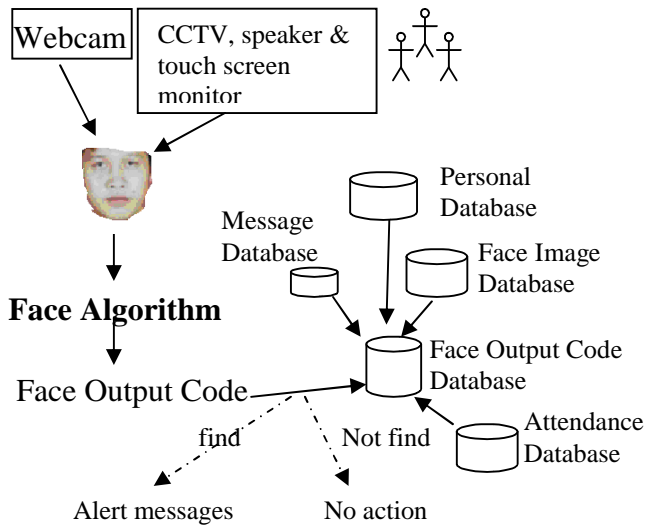

Figure 3. Transaction with Face on Office automation

Message database will be recorded with some personal message like meeting, appointment, email and etc. For privacy reason human can arrange the message alert by setting what kind of personal messages can be alert to them or even there is no message alert, but for organization purposes, the message for meeting can't be denied as a message. Personal database will be recorded with personal employee data like address, phone number, salary, recorded activity and etc.

For attendance system can be also done with human face and not with other biometric recognition. Attendance system with face recognition can automatically recognize human attendance for office, class or every system which record human attendances. With human face as recognizing for attendance, human will have never can cheat their attendance by asking other human to clock their attendance or record the wrong time. Attendance database will be prepared for record every attendance activities.

For office or working which need surveillance for their employee, then human face also can be used to record employee activities fairly. Employer can be helped by system automatically to value their employee without human interfere. The System with human face can capture employee daily activities, what time they come, what time they go home, how much time they do their lunch, how much time they do their toilet activities, how much time they stay in their chair or rooms, how much time they go out from their rooms or how much time they use their computer or tools ? And many questions can be answered as an indicator to survey their employee automatically by system.

Human face will be captured and recognized with webcam or CCTV which equipped with speaker and touch screen monitor. Computer will transform human face with face algorithm become face output code. The face output code which is simple text will be searched on face output code database which is filled with simple code text which represent for each human face as primary key or single identity.

If searching find the match record then the message database will be searched with simple text code primary key and if find then check if the date for the message still up to date and if still up to date then message will be alerted. If can't find there is no action. When employee use the computer with webcam then the alert can be shown with sound if they activate the sound card or with preview message on computer screen. Human when they use their computer even surfing on internet either using internet from office or outside office can be recognized and alerted.

CCTV which equipped with speaker and touch screen will be erected in every strategic place where CCTV can capture and recognize human face easily. When CCTV capture human face, the computer will transform human face with face algorithm become face output code which will be searched on face output code database. If find then with the same face output code will be searched on message database, if find and the date for the message still activate then the message will be alerted either with voices and displaying on computer screen.

If the target person interest with the message alert, they can come to the near CCTV which equipped with speaker and touch screen and look for the message. For looking at data other employee can be done by searching the face output code in personal database and read all the data needed.

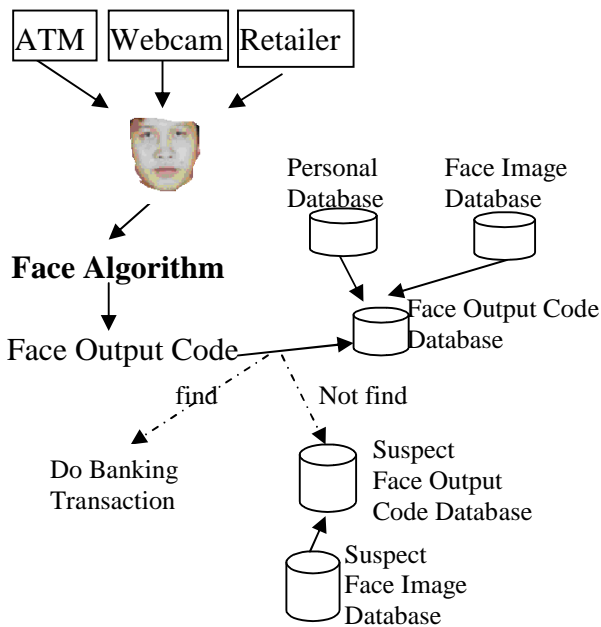

Figure 4. Transaction with Face on Banking system

For banking system, the security will improve significantly. The Banking System can more secure to protect their customer's money and against money laundering or other financial crime. Other banking transaction like transaction by phone and television will be changed as the requirement for face recognition in banking transaction.

The same with implementation in office automation there will be face image database, face output code database and personal database but without message database. If possible message can be used as an improvement for customer personal message alert. The scenario for human face identification and recognizing in office automation will be implemented as well as in banking system. If human face can be recognized then they are recognized as the banking customer, but if can't recognized and insist for doing fraud the face output code will be recorded in suspect face output code database. Suspect image database as complement for face output code database to save image human face.

Every activity by unrecognized human face will be saved automatically by cam or webcam for evidence purpose. Their cam activity can be analyzed for interrogation purpose. In the future for bank security this Suspect Face Output Code Database and Suspect Face Image Database will be searched on world human face centre database for recognizing. If their data can be found in Centre database human face then their criminal activity record will be added. If they can't be found then the new record will be added with criminal activity record adding.

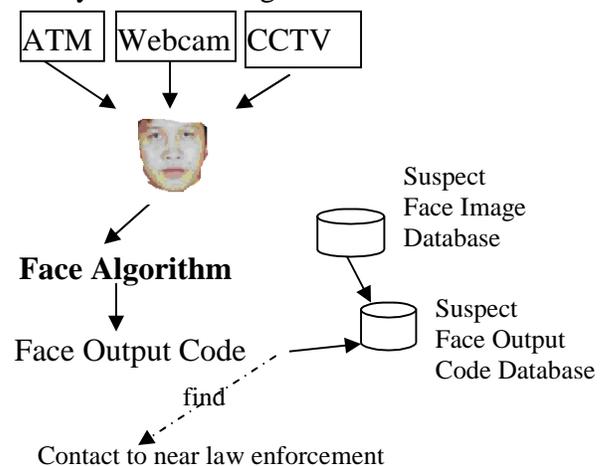

Figure 5. Surveillance for suspect human face

Searching for suspect human face can be done on every webcam access and every CCTV which equipped for surveillance. The concept will be the same with office automation or banking system, where the CCTV or cam will catch human face and transform with face algorithm become face output code. The face output code will be searched on suspect face output code database and if find then face in suspect face image database will be retrieved and automatically system will alert the near law enforcement.

Another face recognition value added can be extended such as when the human face is detected, computer can recognize the human emotion [3][4][5],and human health which combine with speech recognition or body temperature censor [4][6]. Computer can give a joke, bonus or good message when human in stress condition. For some human with disease identification, computer can be reacted as like their doctor to ask them to eat their pill or find their doctor, giving advice such as have a good sleeping and have a nice food. Even the computer will give alert to medical institution as a warning for preparing emergency medical condition.

## 5. Conclusion

Human face as a recognition for human can be powerful to use as a single identity in the world, where the world security will be improved in order to combat terrorism and crime globally. Man can only has one identity and never have more than one identity and man will use their face as their identity as the real recognition in daily human living. Human can be easy recognized globally and system around the world can be easily to recognize human without many screening and wasting time.

Human face from time to time and from generation to generation will be recorded in a huge and large world centre human face database. Real accurate human data in the world will be delivered easily. Other data and researches can be extended for human welfare and security.

Human no need their card anymore as their identification or for transactions. Human can be easily recognized by the system with their face as the system can easily to recognize human faces.

Fighting for terrorism and crime can be delivered by surveillance and searching suspect human face with suspect human face database.